\documentclass[10pt,twocolumn,letterpaper]{article}

\usepackage[pagenumbers]{cvpr} 

\usepackage{graphicx}
\usepackage{amsmath}
\usepackage{amssymb}
\usepackage{booktabs}

\usepackage{enumitem}
\usepackage{multirow}
\usepackage{bbding}
\usepackage[dvipsnames]{xcolor}
\usepackage{indentfirst}
\usepackage[accsupp]{axessibility} 
\usepackage[pagebackref,breaklinks,colorlinks,citecolor=RoyalBlue,]{hyperref}

\usepackage[capitalize]{cleveref}
\crefname{section}{Sec.}{Secs.}
\Crefname{section}{Section}{Sections}
\Crefname{table}{Table}{Tables}
\crefname{table}{Tab.}{Tabs.}

\def\cvprPaperID{6062} 
\def\confName{CVPR}
\def\confYear{2022}

\begin{document}

\title{A Dual Weighting Label Assignment Scheme for Object Detection}

\author{Shuai Li, \quad Chenhang He, \quad Ruihuang Li, \quad Lei Zhang \\
{The Hong Kong Polytechnic University}\\
{\tt\small \{csshuaili, csche, csrhli, cslzhang\}@comp.polyu.edu.hk}}

\maketitle

\begin{abstract}
    Label assignment (LA), which aims to assign each training sample a positive (pos) and a negative (neg) loss weight, plays an important role in object detection. Existing LA methods mostly focus on the design of pos weighting function, while the neg weight is directly derived from the pos weight. Such a mechanism limits the learning capacity of detectors. In this paper, we explore a new weighting paradigm, termed  dual weighting (DW), to specify pos and neg weights separately. We first identify the key influential factors of pos/neg weights by analyzing the evaluation metrics in object detection, and then design the pos and neg weighting functions based on them. Specifically, the pos weight of a sample is determined by the consistency degree between its classification and localization scores, while the neg weight is decomposed into two terms: the probability that it is a neg sample and its importance conditioned on being a neg sample.  Such a weighting strategy offers greater flexibility to distinguish between important and less important samples, resulting in a more effective object detector. Equipped with the proposed DW method, a single FCOS-ResNet-50 detector can reach 41.5\% mAP on COCO under 1$\times$ schedule, outperforming other existing LA methods. It consistently improves the baselines on COCO by a large margin under various backbones without bells and whistles. Code is available at \url{https://github.com/strongwolf/DW}.
\end{abstract}
\section{Introduction}
\begin{figure}[tbp]
    \centering
    \includegraphics[width=0.45\textwidth]{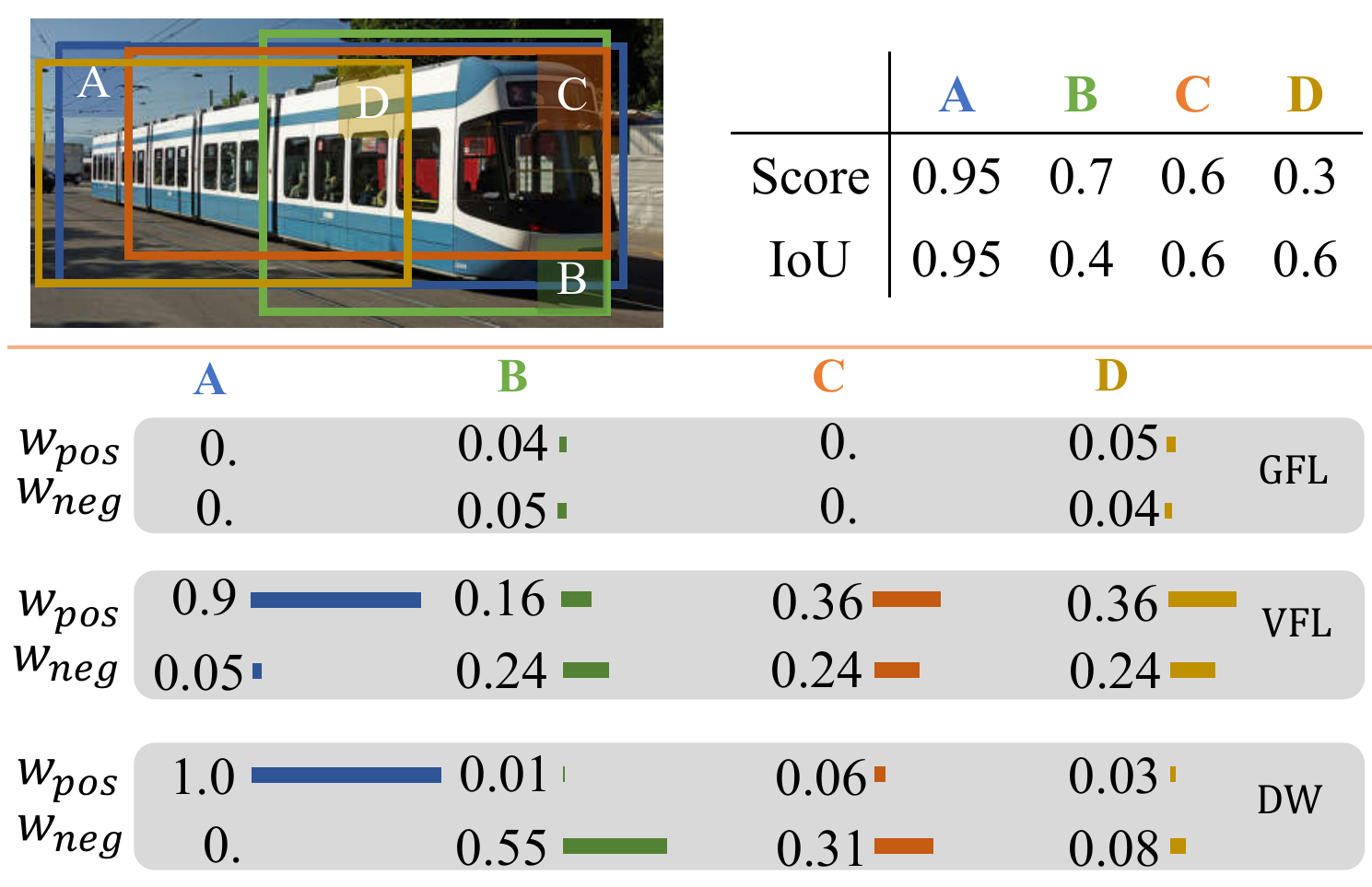}
    \caption{An illustration of the difference between the proposed DW method and existing label assignment methods, \eg, GFL~\cite{gfocal} and VFL~\cite{varifocalnet}. For ambiguous anchors B, C and D, GFL and VFL will assign nearly the same \textit{pos} and \textit{neg} weights to \{B, D\} and \{C, D\}, respectively. In contrast, our DW assigns a distinct (\textit{pos}, \textit{neg}) pair for each anchor.}
    \label{fig1}
\end{figure}
As a fundamental vision task, object detection has been drawing significant attention from researchers for decades. The community has recently witnessed a fast evolution of detectors with the development of convolutional neural networks (CNNs)~\cite{resnet,vgg, densenet,mobilenets,googlenet1,googlenet2,googlenet3} and visual transformers (ViTs)~\cite{transformer1,transformer2,transformer3,detr, deformdetr,conditionaldetr,vit,fastdetr,dynamicdetr}. Current state-of-the-art detectors~\cite{fcos,focalloss,ssd,borderdet,yolo,yolov3,yolov4,featureselect,softanchor,objects_as_points} mostly perform dense detection by predicting class labels and regression offsets with a set of pre-defined anchors. As the basic unit in detector training, anchors need to be assigned with proper classification (\textit{cls}) and regression (\textit{reg}) labels to supervise the training process. Such a label assignment (LA) process can be regarded as a task of assigning loss weight to each anchor. The \textit{cls} loss (\textit{reg} loss can be similarly defined) for an anchor can be generally expressed as:
\begin{equation}
\label{eq1}
\mathcal{L}_{cls}= - w_{pos} \times \ln{(s)} - w_{neg} \times \ln{(1-s)},
\end{equation}
where $w_{pos}$ and $w_{neg}$ are the positive (\textit{pos}) and negative (\textit{neg}) weights, respectively, and $s$ is the predicted classification score. Depending on the design of $w_{pos}$ and $w_{neg}$, the LA methods can be roughly divided into two categories: hard LA and soft LA.

Hard LA assumes that each anchor is either \textit{pos} or \textit{neg}, which means that $w_{pos}$, $w_{neg} \in \{0,1\}$ and $w_{neg}+w_{pos}=1$. The core idea of this strategy is to find a proper division boundary to split the anchors into a positive set and a negative set. The division rule along this line of research can be further categorized into static and dynamic ones. Static rules~\cite{fasterrcnn,ssd,fcos,foveabox} adopt pre-defined metrics such as the IoU or the distance from the anchor center to the ground truth (GT) center to match objects or background to each anchor. Such static assignment rules ignore the fact that
the division boundaries of objects with different sizes and shapes may vary. Recently, many dynamic assignment rules~\cite{iqdet,ota} have been proposed. For instance, ATSS~\cite{atss} splits the training anchors of an object based on their IoU distributions.
Prediction-aware assignment strategies~\cite{detr,paa,noisyanchor} regard the predicted confidence score  as a reliable indicator for estimating an anchor’s quality. Both static and dynamic assignment methods ignore the fact that samples are not equally important. The evaluation metric in object detection suggests that an optimal prediction should have not only a high classification score but also an accurate localization, which implies that anchors with higher consistencies between the \textit{cls} head and \textit{reg} head should have greater importance during training. 

With the above motivation, researchers have opted to assign soft weights to anchors.  GFL~\cite{gfocal} and VFL~\cite{varifocalnet} are two typical methods which define soft label targets based on IoUs and then translate them into loss weights by multiplying a modulation factor. Some other works~\cite{tood,musu} compute sample weights by jointly considering the \textit{reg} score and \textit{cls} score. 
Existing methods mainly focus on the design of \textit{pos} weighting function while the \textit{neg} weight is simply derived from the \textit{pos} weight, which may limit the learning capacity of detectors due to little new supervision information provided by \textit{neg} weights.
We argue that such coupled weighting mechanism cannot distinguish each training sample at a finer level. Fig.~\ref{fig1} shows an example. Four anchors have different prediction results. However, GFL and VFL assign nearly the same (\textit{pos}, \textit{neg}) weight pair to (B, D) and (C, D), respectively. GFL also assigns both zero \textit{pos} and \textit{neg} weight to anchor A and C since each one has the same \textit{cls} score and IoU. 
As the \textit{neg} weighting function  is highly correlated with the \textit{pos} one in existing soft LA methods, anchors  with different attributes can sometimes be assigned nearly the same (\textit{pos}, \textit{neg}) weights, which may impair the effectiveness of the trained detector.

To provide more discriminative supervision signals to the detector, we propose a new LA scheme, termed dual weighting (DW), to specify \textit{pos} and \textit{neg} weights from different perspectives and make them complementary to each other. Specifically, the \textit{pos} weights are dynamically determined by the combination of confidence scores (obtained from the \textit{cls} head) and the \textit{reg} scores  (obtained from the \textit{reg} head). The \textit{neg} weight for each anchor is decomposed into two terms: the probability that it is a \textit{neg} sample and its importance conditioned on being a \textit{neg} sample. The \textit{pos} weight reflects the consistency degree between the \textit{cls} head and \textit{reg} head, and it will push anchors with higher consistencies to move forward in the anchor list, while the \textit{neg} weight reflects the inconsistency degree and pushes the inconsistent anchors to the rear of the list. By this means, at inference the bounding boxes with higher \textit{cls} scores and more precise locations will have better chances to survive after NMS, and those bounding boxes with imprecise locations will fall behind and be filtered out. Referring to Fig.~\ref{fig1}, DW distinguishes four different anchors by assigning them distinct (\textit{pos}, \textit{neg}) weight pairs, which can provide the detector with more fine-grained supervision training signals.

In order to provide our weighing functions with more accurate \textit{reg} scores, we further propose a box refinement operation. Specifically, we devise a learned prediction module to generate four boundary locations based on the coarse regression map, and then aggregate the prediction results of them to get the updated bounding box for the current anchor. This light-weight module enables us to provide more accurate \textit{reg} scores to DW by only introducing moderate computational overhead.

The advantage of our proposed DW method is demonstrated by comprehensive experiments on MS COCO~\cite{coco}. In particular, it boosts the FCOS~\cite{fcos} detector with ResNet-50~\cite{resnet} backbone to a 41.5/42.2 AP w/wo box refinement on the COCO validation set under the common 1$\times$ training scheme, surpassing other LA methods.

\section{Related Work}

\textbf{Hard Label Assignment.} Labeling each anchor to be a \textit{pos} or \textit{neg} sample is a key procedure to train a detector. Classical anchor-based object detectors~\cite{fasterrcnn,ssd} set an anchor’s label by measuring its IoU with the GT objects. 
Recently, anchor-free detectors have attracted much attention due to their concise design and comparable performance. 
FCOS~\cite{fcos} and Foveabox~\cite{foveabox} both select \textit{pos} samples by a center sampling strategy: anchors that are close to the GT centers are sampled as positives and others are negatives or ignored during training. The above mentioned LA methods adopt a fixed rule for GT boxes with diverse shapes and sizes, which is sub-optimal.

Some advanced LA strategies~\cite{atss,paa,ota,multiple_anchor,liu2020hambox,ming2020dynamic} have been proposed to dynamically choose \textit{pos} samples for each GT. ATSS~\cite{atss} selects top-k anchors from each level of the feature pyramid and adopts the \texttt{mean+std} IoU of these top anchors as the \textit{pos}/\textit{neg} division threshold. PAA~\cite{paa} adaptively separates anchors into \textit{pos}/\textit{neg} ones based on the joint status of \textit{cls} and \textit{reg} losses in a probabilistic manner. OTA~\cite{ota} handles the LA problem from a global perspective by formulating the assignment process as an optimal transportation problem. Transformer-based detectors~\cite{detr,conditionaldetr,deformdetr,dynamicdetr} adopt a one-to-one assignment scheme by finding the best \textit{pos} sample for each GT. Hard LA treats all samples equally, which however is less compatible with the evaluation metric in object detection.

\textbf{Soft Label Assignment.} Since the predicted boxes have different qualities in evaluation, the samples should be treated differently during training. Many works~\cite{primesample,focalloss,varifocalnet,gfocal,gfocalv2} have been proposed to address the inequality issue of training samples. Focal Loss~\cite{focalloss} adds a modulated factor on the cross entropy loss to down-weight the loss assigned to well-classified samples, which pushes the detector to focus on hard samples. Generalized focal loss~\cite{gfocal} assigns each anchor a soft weight by jointly considering the \textit{cls} score and localization quality. Varifocal loss~\cite{varifocalnet} utilizes an IoU-aware \textit{cls} label to train the \textit{cls} head. Most methods mentioned above focus on computing the \textit{pos} weight and simply define the \textit{neg} weight as a function of 1 - $w_{pos}$. In this paper, we decouple this procedure and separately assign \textit{pos} and \textit{neg} loss weights for each anchor. 
Most soft LA methods assign weights to loss. There is a special case that weights are assigned to score, which can be formulated as $\mathcal{L}_{cls} = -\ln{( w_{pos} \times s) -\ln{ ( 1 - w_{neg} \times s)}}$. Typical methods include FreeAnchor~\cite{freeanchor} and Autoassign~\cite{autoassign}. It should be noted that our method is distinct from them. To match anchors in a fully differential manner, $w_{pos}$ and $w_{neg}$ in Autoassign still receive gradients. In our method, however, the loss weights are carefully designed and entirely detached from the network, which is a common practice for weighting loss.
\section{Proposed Method}
\subsection{Motivation and Framework}
To be compatible with NMS, a good dense detector should be able to predict consistent bounding boxes that have both high classification scores and precise locations. However, if all the training samples are equally treated, there will be a misalignment between the two heads: the location with the highest category score is usually not the best position for regressing the object boundary. This misalignment can degrade the performance of detectors, especially under high IoU metrics. Soft LA, which treats the training samples in a soft manner by weighting loss, is an attempt to enhance the consistency between the \textit{cls} and \textit{reg} heads. With soft LA, the loss of an anchor can be expressed as:
\begin{equation}
\begin{aligned}
\mathcal{L}_{c l s}&=-w_{pos } \times \ln (s)-w_{neg } \times \ln (1-s), \\
\mathcal{L}_{reg }&=w_{reg } \times \ell_{reg }\left(b, b^{\prime}\right),
\end{aligned}
\label{eq2}
\end{equation}
where $s$ is the predicted \textit{cls} score, $b$ and $b^{\prime}$ are the locations of the predicted bounding box and the GT object, respectively, and $\ell_{reg }$ is the regression loss such as Smooth $L_1$ Loss~\cite{fasterrcnn}, IoU Loss~\cite{iouloss} and GIoU Loss~\cite{giou}. The inconsistency problem between \textit{cls} and \textit{reg} heads can be mitigated by assigning larger $w_{pos}$ and $w_{reg}$ to anchors with higher consistencies. These well-trained anchors are thus able to predict high \textit{cls} scores and precise locations simultaneously at inference.

\begin{table}[tb]
\centering
\caption{Comparison of different weighting functions.}
\vspace{-3mm}
\label{table1}
\footnotesize{
\setlength{\tabcolsep}{1.5mm}
\begin{tabular}{c|c|c|c}
\toprule[1.0pt]
Method & $w_{pos}$ & $w_{neg}$ & $t$ \\
\hline
\hline
GFL~\cite{gfocal} & $(s-t)^2 \times t$ & $(s-t)^2 \times (1-t)$ & $IoU$ \\
VFL~\cite{varifocalnet} & $t \times t$ & $t \times (1-t)$ & $IoU$ \\
TOOD~\cite{tood} & $(s-t)^2 \times t$ & $(s-t)^2 \times (1-t)$ & $f(IoU, s)$ \\
MuSu~\cite{musu} & $(s-t)^2 \times t$ & $s^2 \times (1-t)^4$ & $f(IoU, s)$ \\
Ours (DW)  & $f_{pos}(IoU,s)$ & $P_{neg} \times I_{neg}$ & - \\
\bottomrule[1.0pt]
\end{tabular}
}
\vspace{-4mm}
\end{table}
\begin{figure*}[tbp]
    \centering
    \includegraphics[width=0.9\textwidth]{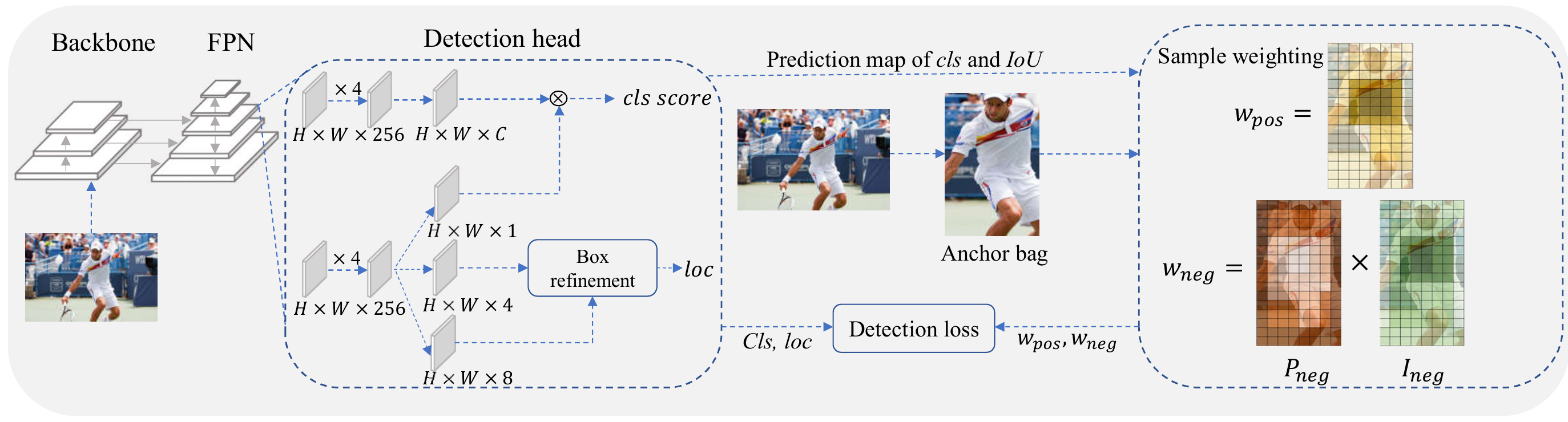}
    \caption{Pipeline of DW. The left part shows the overall detection model which consists of backbone, FPN and detection head. The outputs from the classification branch ($H \times W \times C$) and the centerness branch ($H \times W \times 1$) are multiplied as the final \textit{cls} score. The box refinement module utilizes the four predicted boundary points ($H \times W \times 8$) to adjust the coarse prediction ($H \times W \times 4$) to finer locations.The right part shows the weighting process. Given an object, a candidate anchor bag is first constructed by selecting the anchor points near the object center. Each anchor will then be assigned a \textit{pos} weight and \textit{neg} weight from different aspects.}
    \label{fig2}
\end{figure*}

Existing works commonly set $w_{reg}$ equal to $w_{pos}$ and mainly focus on how to define the consistency and integrate it into loss weights. Table~\ref{table1} summarizes the formulations of $w_{pos}$ and $w_{neg}$  for  a \textit{pos} anchor in recent representative methods. One can see that current methods commonly define a metric \textit{t} to indicate the consistency degree between the two heads at the anchor level, and then design the inconsistency metric as a function of $1-t$. The consistent and inconsistent metrics are finally integrated into the \textit{pos} and \textit{neg} loss weights by adding a scaling factor ($(s-t)^2$, $s^2$ or $t$), respectively.

Different from above methods where $w_{pos}$ and $w_{neg}$ are highly correlated, we propose to set \textit{pos} and \textit{neg} weights separately in a prediction-aware manner. Specifically, the \textit{pos} weighting function takes the predicted \textit{cls} score $s$  and the IoU between the predicted box and the GT object as inputs, and set the \textit{pos} weight by estimating the consistency degree between the \textit{cls} and \textit{reg} heads. The \textit{neg} weighting function takes the same inputs as the \textit{pos} weighting function but formulates the \textit{neg} weight as the multiplication of two terms: the probability that the anchor is a \textit{neg} one, and its importance conditioned on that it is \textit{neg}. By this way, the ambiguous anchors that have similar \textit{pos} weights can receive more fine-grained supervision signals with distinct \textit{neg} weights, which is not available in existing methods.

The pipeline of our DW framework is shown in Fig.~\ref{fig2}. As a common practice~\cite{ota,musu,tood,fcos}, we first construct a bag of candidate positives for each GT object by selecting anchors near the GT center (center prior). Anchors outside the candidate bag are considered as \textit{neg} samples which will not be involved in the design process of weighting functions since their statistics (\eg, IoU, \textit{cls} score) are very noisy at the early training stage. Anchors inside the candidate bag will be assigned to three weights including $w_{pos}$, $w_{neg}$ and $w_{reg}$, to supervise the training process more effectively.

\subsection{Positive Weighting Function}
The \textit{pos} weight of a sample should reflect its importance for accurately detecting an object in both classification and localization. We try to find out the factors affecting this importance by analyzing the evaluation metric of object detection. During testing on COCO, all the predictions for one category should be properly ranked by a ranking metric. Existing methods commonly use the \textit{cls} score~\cite{fasterrcnn} or the combination of \textit{cls} score and predicted IoU~\cite{atss} as the ranking metric. The correctness of each bounding box will be checked from the begin of the ranking list. A prediction will be defined as a correct one if and only if:
\renewcommand{\theenumi}{\alph{enumi}}
\begin{enumerate}
\item 
\label{item_a}
The IoU between the predicted bounding box and its nearest GT object is larger than a threshold $\theta$;
\item
No box that satisfies the above condition ranks in front of the current box.
\end{enumerate}

In a word, only the first bounding box that has a larger IoU than $\theta$ in the prediction list will be defined as a \textit{pos} detection, while all the other bounding boxes should be considered as false positives of the same GT. It can be seen that high ranking score and high IoU are both \textit{sufficient} and \textit{necessary} conditions for a \textit{pos} prediction. This implies that anchors that simultaneously satisfy the two conditions are more likely to be defined as \textit{pos} predictions during testing, and thus they should have higher importance during training. From this perspective, the \textit{pos} weight $w_{pos}$ should be positively correlated with the IoU and ranking score, \ie, $w_{pos}\propto IoU$ and $w_{pos} \propto s$. To specify the \textit{pos} function, we first define a consistency metric, denoted as $t$, to measure the alignment degree between the two conditions:
\begin{equation}
\begin{aligned}
t = s \times IoU^ \beta,\\
\end{aligned}
\label{eq3}
\end{equation}
where $\beta$ is used to balance the two conditions. To encourage a large variance of \textit{pos} weights among different anchors, we add an exponential modulation factor:
\begin{equation}
\begin{aligned}
w_{pos} = e^ {\mu t} \times t,\\
\end{aligned}
\label{eq4}
\end{equation}
where $\mu$ is a hyper-parameter to control the relative gaps of different \textit{pos} weights. Finally, the \textit{pos} weight of each anchor  for each instance is normalized by the sum of all \textit{pos} weights within the candidate bag.

\subsection{Negative Weighting Function}
Though \textit{pos} weights can enforce consistent anchors to have both high \textit{cls} scores and large IoUs, the importance of less consistent anchors cannot be distinguished by \textit{pos} weights. 
Referring to Fig.~\ref{fig1}, anchor D has a finer location (a larger IoU than $\theta$ ) but a lower \textit{cls} score, while anchor B has a coarser location (a smaller IoU than $\theta$) but a higher \textit{cls} score. They may have the same consistency degree $t$ and thus will be pushed forward with the same \textit{pos} strength which can not reflect their differences.
To provide more discriminative supervision information for the detector, we propose to indicate their importance faithfully by assigning more distinct \textit{neg} weights to them, which are defined as the multiplication of the following two terms.

\textbf{Probability of being a Negative Sample.}  According to the evaluation metric of COCO, an IoU smaller than $\theta$ is a sufficient condition for a false prediction. This means that a predicted bounding box that does not satisfy the IoU metric will be regarded as a \textit{neg} detection, even if it has a high \textit{cls} score. That is, IoU is the only factor to determine the probability of being a \textit{neg} sample, denoted by $P_{neg}$. Since COCO adopts an IoU interval ranging from 0.5 to 0.95 to estimate AP, the probability $P_{neg}$ for a bounding box should satisfy the following rules:
\begin{equation}
    P_{neg} = \begin{cases}
    1, & \textit{if   } \; \text{IoU $<$ 0.5}, \\
    [0,1], & \textit{if   } \; \text{IoU $\in$ [0.5,0.95]}, \\
    0, & \textit{if   } \; \text{IoU $>$ 0.95},
    \end{cases}
\label{eq5}
\end{equation}

\begin{figure}[tbp]
    \centering
    \includegraphics[width=0.35\textwidth]{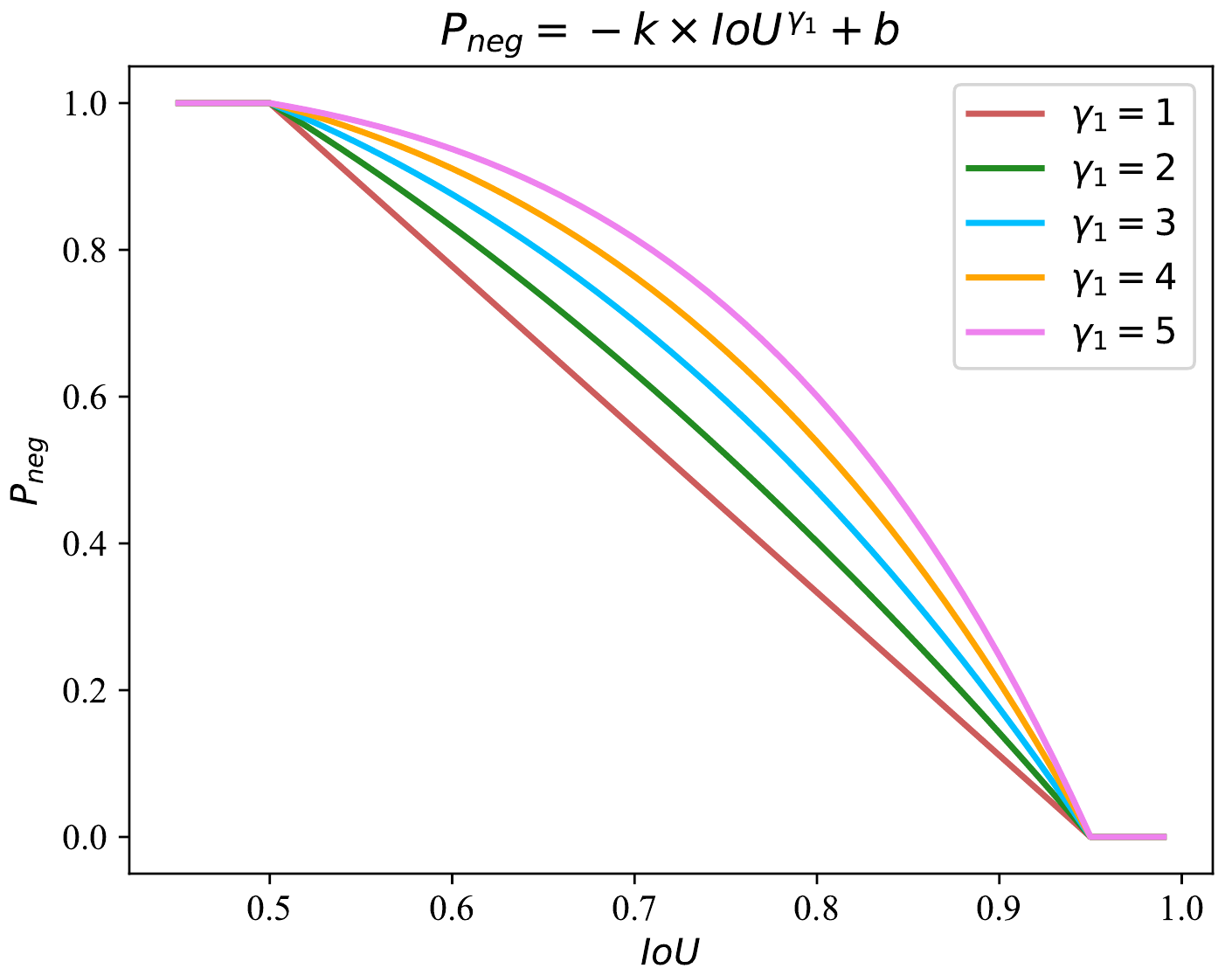}
    \caption{Curves of $P_{neg}$ in [0.5,0.95] vs. IoU with different $\gamma_1$.}
    \vspace{-3mm}
    \label{neg_curve}
\end{figure}

 Any monotonically decreasing function defined within the interval [0.5,0.95] is qualified for $P_{neg}$. For simplicity, we instantiate $P_{neg}$ as the following function:
\begin{equation}
P_{neg} = -k \times IoU ^ {\gamma_1} + b, \quad \textit{if  } \text{ IoU $\in$ [0.5,0.95]}, \\
\label{eq5}
\end{equation}
which passes through the points (0.5, 1) and (0.95,0). Once $\gamma_1$  is determined, the parameters  $k$ and $b$ can be obtained by the method of undetermined coefficients. Fig.~\ref{neg_curve} plots the curves of $P_{neg}$ vs. IoU with different values of $\gamma_1$.

\textbf{Importance Conditioned on being a Negative Sample.} At inference, a \textit{neg} prediction in the ranking list will not affect the recall but decrease the precision. To delay this process, the \textit{neg} bounding boxes should rank as behind as possible, \ie, their ranking scores should be as small as possible. Based on this point, the \textit{neg} predictions with larger ranking scores are more important than those with smaller ranking scores as they are harder examples for network optimization. Thus, the importance of \textit{neg} samples, denoted by $I_{neg}$, should be a function of the ranking score. For simplicity, we set it as 
\begin{equation}
I_{neg} = s^{\gamma_2}, \\
\label{eq6}
\end{equation}
where $\gamma_2$ is a factor to indicate how much preference should be given to important \textit{neg} samples.

Finally, the \textit{neg} weight $w_{neg}=P_{neg} \times I_{neg}$ becomes
\begin{equation}
\small{
w_{neg}= 
    \begin{cases}
    s^{\gamma_2}, &  \textit{if  } \; \text{IoU $<$ 0.5}, \\
    (-k \times IoU^{\gamma_{1}}+b) \times s^{\gamma_{2}}, &  \textit{if  } \; \text{IoU $\in$ [0.5,0.95]}, \\
    0, & \textit{if  } \; \text{IoU $>$ 0.95},
    \end{cases}
    \label{eq7}
      }
\end{equation}
which is negatively correlated with IoU but positively correlated with $s$. It can be seen that for two anchors with the same \textit{pos} weight, the anchor with a smaller IoU will have a larger \textit{neg} weight. The definition of $w_{neg}$ is well compatible with the inference process and it can further distinguish ambiguous anchors having almost the same \textit{pos} weights. Please refer to Fig.~\ref{fig1} for examples.

\subsection{Box Refinement}
\begin{figure}[tbp]
    \centering
    \includegraphics[width=0.3\textwidth]{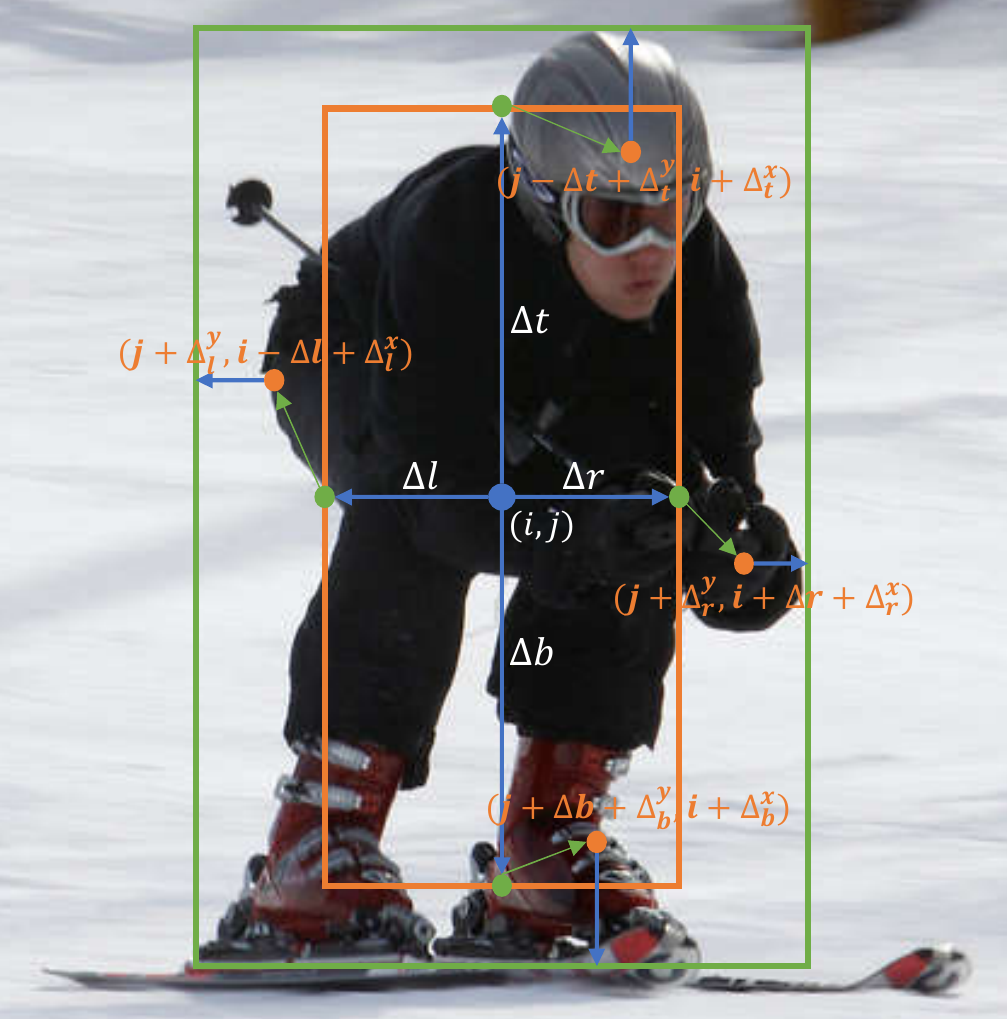}
    \caption{Illustration of the box refinement operation. A coarse bounding box (orange box) of an anchor at location (\textit{j,i}) is first generated by predicting four distances = $\{\Delta l, \Delta t, \Delta r, \Delta b\}$. Four boundary points (orange points) are then predicted with respect to the four side points (green points). Finally, a finer bounding box (green box) is generated by aggregating the prediction results of the four boundary points.}
    \vspace{-4mm}
    \label{box_refinement}
\end{figure}
Since both the \textit{pos} and \textit{neg} weighting functions take IoU as an input, more accurate IoUs can induce higher quality samples, benefiting the learning of stronger features. We propose a box refinement operation to refine the bounding boxes based on the predicted offset map $O \in R^ {H \times W \times 4}$, where $O(j,i) = \{\Delta l, \Delta t, \Delta r, \Delta b\}$ represents the predicted distances from the center of current anchor to the leftmost \textit{l}, topmost \textit{t}, rightmost \textit{r} and bottommost \textit{b} sides of the GT object, respectively, as shown in Fig.~\ref{box_refinement}. Motivated by the fact that points near the object boundary are more likely to predict accurate locations, we devise a learnable prediction module to generate a boundary point for each side based on the coarse bounding box. Referring to Fig.~\ref{box_refinement}, the coordinates of the four boundary points are defined as:
\begin{equation}
\footnotesize{
\begin{aligned}
B_{l}&=\left(j+\Delta_{l}^{y}, i-\Delta l+\Delta_{l}^{x}\right), B_{t}=\left(j-\Delta t+\Delta_{t}^{y}, i+\Delta_{t}^{x}\right), \\
B_{r}&=\left(j+\Delta_{r}^{y}, i+\Delta r +\Delta_{r}^{x}\right), B_{b}=\left(j+\Delta b+\Delta_{b}^{y}, i+\Delta_{b}^{x}\right),
\end{aligned}
\label{eq8}
    }
\end{equation}
\noindent where $\left\{\Delta_{l}^{x}, \Delta_{l}^{y}, \Delta_{t}^{x}, \Delta_{t}^{y}, \Delta_{r}^{x}, \Delta_{r}^{y}, \Delta_{b}^{x}, \Delta_{b}^{y}\right\}$ are the outputs of the refinement module.

The refined offset map $O^\prime$ is updated as:
\begin{equation}
\footnotesize{
O^{\prime}(j, i)=\left\{\hspace{-1.mm}\begin{array}{l}
\Delta l+\Delta_{l}^{x}+O(B_{l},0), \; \Delta t+\Delta_{t}^{y}+O(B_{t},1) \\
\Delta r+\Delta_{r}^{x}+O(B_{r},2), \; \Delta b+\Delta_{b}^{y}+O(B_{b},3)
\end{array}\hspace{-1.mm}\right\}
\label{eq10}
    }
\end{equation}

\subsection{Loss Function}
The proposed DW scheme can be applied to most existing dense detectors. Here we adopt the representative dense detector FCOS~\cite{fcos} to implement DW. As shown in Fig.~\ref{fig2}, the whole network structure comprises a backbone, FPN and detection head. Following the conventions~\cite{musu,fcos,autoassign}, we multiply the outputs of centerness branch and classification branch as the final \textit{cls} score. The final loss of our network is 
\begin{equation}
    \mathcal{L}_{det}=\mathcal{L}_{cls}+\beta \mathcal{L}_{reg},
\end{equation}
where $\beta$ is a balancing factor which is the same as the one in Eq.~\ref{eq3}, and 
\begin{equation}
\small{
\begin{aligned}
\mathcal{L}_{c l s}&=\sum\nolimits_{n=1}^{N} -w_{p o s}^{n} \times \ln \left(s^{n}\right)-w_{n e g}^{n} \times \ln \left(1-s^{n}\right) \\
&+\sum\nolimits_{m=1}^{M} F L\left(s^{m}, 0\right), \\
\mathcal{L}_{reg}&=\sum\nolimits_{n=1}^{N} w_{pos}^{n}  \times GIoU\left(b, b^{\prime}\right),
\end{aligned}
    }
\end{equation}
\noindent 
where $N$ and $M$ are the total number of anchors inside and outside the candidate bag, respectively, FL is the Focal Loss~\cite{focalloss}, GIoU is the regression loss~\cite{giou}, $s$ is the predicted \textit{cls} score, $b$ and $b^{'}$ are the locations of the predicted box and the GT object, respectively.
\section{Experiments}
\textbf{Dataset and Evaluation Metric.} Extensive experiments are conducted on the large-scale detection benchmark MS-COCO~\cite{coco} which contains 115K, 5K and 20K images for \texttt{train}, \texttt{val} and \texttt{test-dev} sets, respectively. We report the analysis and ablation studies on \texttt{val} set and compare with other state-of-the-arts on the \texttt{test-dev} set. The performance is measured by COCO Average Precision (AP).

\textbf{Implementation Details.} We use ResNet-50 pretrained on ImageNet~\cite{imagenet} with FPN~\cite{fasterrcnn} as our backbone for all experiments unless otherwise specified. Following the common practice, most models are trained with 12 epochs which is denoted as 1$\times$ in~\cite{mmdetection}. The initial learning rate is 0.01 and is decayed by a factor of 10 after the $8^{th}$ and $11^{th}$ epoch. For all ablations, we use an image scale of 800 pixels for training and testing unless otherwise specified. All experiments are trained with SGDM~\cite{sgd} on 8 GPUs with a total batch size 16 (2 images per GPU). At inference, we filter out background boxes with a threshold 0.05 and remove redundant boxes by NMS with a threshold 0.6 to get the final predicted results. The hyper-parameters $\gamma_1$, $\gamma_2$, $\beta$ and $\mu$ are set to 2, 2, 5 and 5, respectively. 

\begin{table}[tb]
\centering
\caption{Detection performances by setting different hyper-parameters in $w_{pos}$.}
\vspace{-2mm}
\label{pos_hyper}
\scalebox{0.8}{
\setlength{\tabcolsep}{1.2mm}
\begin{tabular}{c|cccccc|cccc}
\toprule[1pt]
$\beta$ & \multicolumn{6}{c|}{5}           & 3    & 4    & 6    & 7       \\
\hline
$\mu$   & 3    & 4    & 5    & 6    & 7    & 8    & \multicolumn{4}{c}{5}  \\
\hline
\hline
AP      & 40.8 & 41.2 & \textbf{41.5} & \textbf{41.5} & 41.4 & 41.2 & 40.8 & 41.3 & 41.4 & 41      \\
AP50    & 59.1 & 59.7 & 59.8 & 60.1 & 59.8 & 59.6 & 59.9 & 59.9 & 59.6 & 59      \\
AP75    & 43.9 & 44.2 & 45   & 44.6 & 45.1 & 44.5 & 43.6 & 44.5 & 44.9 & 44.4   \\
\bottomrule[1pt]
\end{tabular}}


\end{table}
\begin{table}[tb]
\centering
\caption{Detection performances by setting different values of $\gamma_1$ and $\gamma_2$ in $w_{neg}$.}
\vspace{-2mm}
\label{neg_hyper}
\scalebox{0.8}{
\setlength{\tabcolsep}{3.5mm}
\begin{tabular}{c|c|ccc}
\toprule[1pt]
$\gamma_1$ & $\gamma_2$ & AP & AP50 & AP75 \\
\hline
\hline

 1 & 1 & 41. & 59.2 & 44.1 \\
 1 & 2 & 41.3 & 59.7 & 44.6 \\
 2 & 2 & \textbf{41.5} & 59.8 & 45 \\
 3 & 2 & 41.3 & 59.7 & 44.4 \\
 4 & 2 & 41.2 & 59.4 & 44.4 \\
 5 & 2 & 41.1 & 59.5 & 44.5 \\
 2 & 3 & 41.3 & 59.6 & 44.5 \\
\bottomrule[1pt]
\end{tabular}}
\end{table}
\subsection{Ablation Studies}
\textbf{Hyper-parameters of Positive Weighting.} There are two hyper-parameters for \textit{pos} weights: $\beta$ and $\mu$. $\beta$ balances the contributions between the \textit{cls} score and the IoU in the consistency metric $t$. As $\beta$ increases, the contribution degree of IoU also increases. $\mu$ controls the relative scales of \textit{pos} weights. A larger $\mu$ enables the most consistent samples to have relatively larger \textit{pos} weights compared with less consistent samples.  We show the performance of DW by varying $\beta$ from 3 to 7 and $\mu$ from 3 to 8 in Table~\ref{pos_hyper}. One can see that the best result is achieved when $\beta$ is 5 and $\mu$ is 5. Other combinations of $\beta$ and $\mu$ will degrade the AP performance from 0.1 to 0.7. Therefore, we set $\beta$ and $\mu$ to 5 in all the rest experiments.

\textbf{Hyper-parameters of Negative Weighting.} We also conduct several experiments to investigate the robustness of DW to hyper-parameters $\gamma_1$ and $\gamma_2$, which are used to modulate the relative scales of \textit{neg} weights. The AP results by using different combinations of $\gamma_1$ and $\gamma_2$ range from 41 to 41.5, as shown in Table~\ref{neg_hyper}. This implies that the performance of DW is not sensitive to the two hyper-parameters. We adopt $\gamma_1=2$, $\gamma_2=2$ in all our experiments.

\textbf{Construction of Candidate Bag.} As a common practice in object detection, soft LA is only applied on anchors inside the candidate bag. We test three ways of candidate bag construction which are all based on the distance $r$ (normalized by the feature stride) from the anchor point to the corresponding GT center. The first way is to select anchors with a distance smaller than a threshold. The second is to select the top-k nearest anchors from each level of FPN. The third is to give each anchor a soft center weight $e^{-r^2}$ and multiply it with $w_{pos}$. The results are shown in Table~\ref{candidate_bag}. It can be seen that the AP performance fluctuates slightly between 41.1 and 41.5, which demonstrates that our DW is robust to the separation methods of candidate bag.
\begin{table}[tb]
\centering
\caption{Comparisons of different ways to select candidate bag.}
\vspace{-2mm}
\label{candidate_bag}
\scalebox{0.8}{
\begin{tabular}{cc|ccc}
\toprule[1pt]
\multicolumn{2}{c|}{Center prior}  & AP & AP50 & AP75 \\
\hline
\hline
\multirow{5}{*}{Threshold} & 1 & 41.2 & 59.7 & 44.7 \\
 & 1.3 & 41.3 & 59.6 & 44.4 \\
 & 1.7 & 41.4 & 59.5 & 44.6 \\
 & 2.0 & 41.3 & 59.6 & 44.4 \\
 & 2.5 & 41.1 & 59.1 & 44.3 \\
 \hline
  \multirow{3}{*}{Top-k} & 9 & 41.2 & 59.4 & 44.3 \\
 & 12  & 41.2 & 59.4 & 44.6 \\
 & 15  & 41.2 & 59.6 & 44.4 \\
 
 \hline
  Soft center prior & $e^{-r^2}$ & \textbf{41.5} & 59.8 & 45.0 \\
\bottomrule[1pt]
\end{tabular}
}

\end{table}

\begin{table}[tb]
\centering
\caption{Comparisons of different ways to formulate $w_{neg}$.}
\vspace{-2mm}
\label{neg_function}
\scalebox{0.85}{
\begin{tabular}{c|cc|ccc}
\toprule[1pt]
\multirow{2}{*}{$w_{pos}$} & \multicolumn{2}{c|}{$w_{neg}$} & \multirow{2}{*}{AP} & \multirow{2}{*}{AP50} & \multirow{2}{*}{AP75} \\
\cline{2-3}
 & $P_{neg}$ & $I_{neg}$ & & & \\
 \hline
 \hline
 $\surd$ & $\times$ & $\times$ & 39.5 & 58.6 & 42.9 \\
 $\surd$ & $\surd$ & $\times$ & 40.5 & 58.7 & 43.9 \\
  $\surd$ & $\times$ & $\surd$ & 40.0 & 58.5 & 42.9 \\
 $\surd$ & \multicolumn{2}{c|}{$1-w_{pos}$} & 40.7 & 59.5 & 44.1 \\
$\surd$ & $\surd$ & $\surd$ & \textbf{41.5} & 59.8 & 45.0 \\
\bottomrule[1pt]
\end{tabular}
}
\end{table}
\begin{table}[tb]
\centering
\caption{Comparison among different weighting strategies of LA.}
\vspace{-2mm}
\label{comp_la}
\scalebox{0.95}{
\setlength{\tabcolsep}{1.mm}
\begin{tabular}{l|ccc|c}
\toprule[1pt]
Method & AP & AP50 & AP75 & Reference \\
\hline\hline
FoveaBox~\cite{foveabox} & 36.4 & 55.8 & 38.8 &  -\\
 FCOS~\cite{fcos} & 38.6 & 57.4 & 41.4 & ICCV19 \\
ATSS~\cite{atss} & 39.2 & 57.4 & 42.2 & CVPR20 \\
PAA~\cite{paa} & 40.4 & 58.4 & 43.9 & ECCV20 \\
OTA~\cite{ota} & 40.7 & 58.4 & 44.3 & CVPR21 \\
\hline
Autoassign~\cite{autoassign} & 40.4 & 59.6 & 43.7 &- \\
Autoassign(detach) & 39.8 & 59.6 & 42.8 &- \\
Autoassign(weight loss) & 36.6 & 56.2 & 39.1 &- \\
NoisyAnchor~\cite{noisyanchor} & 38.0 & 56.9 & 40.6  & CVPR2020 \\
MAL~\cite{multiple_anchor} & 39.2 & 58.0 & 42.3 & CVPR2020 \\
GFL~\cite{gfocal} & 39.9 & 58.5 & 43.0 & NeurIPS20 \\
VFL~\cite{varifocalnet} & 40.2 & 58.2 & 44.0 & CVPR21 \\
FCOS+GFLv2~\cite{gfocalv2} & 40.6 & 58.2 & 43.9 & CVPR21 \\
ATSS+GFLv2~\cite{gfocalv2} & 41.1 & 58.8 & 44.9 & CVPR21 \\
MuSu~\cite{musu} & 40.6 & 58.9 & 44.3 & ICCV21 \\
TOOD~\cite{tood} & 40.3 & 58.5 & 43.8 & ICCV21 \\
DW & 41.5 & 59.8 & 45.0 &  \\
DW+box refine & \textbf{42.2} & 60.4 & 45.3 &  \\
\bottomrule[1pt]
\end{tabular}
}
\end{table}

\textbf{Design of Negative Weighting Function.}
We investigate the influence of \textit{neg} weighting functions by replacing it with other alternatives, as shown in
Table~\ref{neg_function}. We can see that only using the \textit{pos} weights degrades the performance to 39.5, which indicates that for some low-quality anchors, only assigning them small $w_{pos}$ is not enough to decrease their ranking scores. They can be enforced to rank behind with a larger $w_{neg}$, leading to a higher AP during testing. Without using $I_{neg}$ or $P_{neg}$, we obtain 40.5 AP and 40.0 AP, respectively, which verifies that both the two terms are necessary. As done in existing methods, we attempt to replace $w_{neg}$ with $1-w_{pos}$, but achieve a performance of 40.7 AP, 0.8 points lower than our standard DW.

\textbf{Box Refinement.} Without box refinement, our DW method reaches 41.5 AP, which to our best knowledge is the first method to achieve a performance of more than 41 AP on COCO without increasing any parameters and training cost over FCOS-ResNet-50. With box refinement, DW can reach 42.2 AP, as shown in Table~\ref{comp_la}. Table~\ref{comp_sota} also shows that box refinement can consistently boost the performance of DW with different backbones.

\textbf{Weighting Strategies.} To demonstrate the effectiveness of our DW strategy, we compare it with other LA methods using different weighting strategies. The results are shown in Table~\ref{comp_la}. The first five rows are hard LA methods while the others are soft LA. 

The best performance for hard LA is achieved by OTA, 40.7 AP. Since OTA formulates LA as an Optimal Transport problem, it will increase the training time by more than 20\%. GFLv2 utilizes an extra sophisticated branch to estimate the localization quality and 
 achieves the second best performance of 41.1 AP among soft LA methods.

Unlike the mainstream methods where weights are assigned to loss, Autoassign assigns weights to \textit{cls} score and updates them by their gradients during training. We tried to detach the weights in Autoassign and assigned them to loss, but only obtained 39.8 and 36.6 AP, respectively, 0.6 and 3.8 points lower than the original performance. This implies that the weighting scheme in Autoassign cannot work well when adapting it to the mainstream practice.
\begin{table*}[tb]
\centering
\caption{Performance comparison with state-of-the-art dense detectors on COCO 2017 \texttt{test-dev} set. All models listed below adopt multi-scale training. `*' indicates we use the same detection head proposed in TOOD~\cite{tood}. `Aux.' means the auxiliary learning module.}
\vspace{-2mm}
\label{comp_sota}
\scalebox{0.92}{
\setlength{\tabcolsep}{2.5mm}
\begin{tabular}{l|c|c|ccc|ccc}
\toprule[1pt]
Method     & Backbone   & Aux. & AP   & AP50 & AP75 & AP$_S$  & AP$_M$  & AP$_L$   \\
\hline
\hline
FCOS \cite{fcos}       & ResNet-101  & $\times$    & 41.5 & 60.7 & 45.0 & 24.4 & 44.8 & 51.6  \\
ATSS \cite{atss}       & ResNet-101 & $\times$          & 43.6 & 62.1 & 47.4 & 26.1 & 47.0 & 53.6  \\
PAA \cite{paa}        & ResNet-101  & $\times$        & 44.8 & 63.3 & 48.7 & 26.5 & 48.8 & 56.3  \\
GFL \cite{gfocal}     & ResNet-101 & $\times$           & 45.0 & 63.7 & 48.9 & 27.2 & 48.8 & 54.5  \\
OTA \cite{ota}        & ResNet-101  & $\times$         & 45.3 & 63.5 & 49.3 & 26.9 & 48.8 & 56.1  \\
IQDet \cite{iqdet}      & ResNet-101 & $\times$          & 45.1 & 63.4 & 49.3 & 26.7 & 48.5 & 56.6  \\
MuSu \cite{musu}       & ResNet-101  & $\times$          & 44.8 & 63.2 & 49.1 & 26.2 & 47.9 & 56.4  \\
Autoassign \cite{autoassign} & ResNet-101 & $\times$          & 44.5 & 64.3 & 48.4 & 25.9 & 47.4 & 55.0  \\
VFL \cite{varifocalnet}        & ResNet-101 & $\times$          & 44.9 & 64.1 & 48.9 & \textbf{27.1} & 48.7 & 55.1  \\
DW (ours)  & ResNet-101 & $\times$           & \textbf{46.2} & \textbf{64.8} & \textbf{50.0} & \textbf{27.1} & \textbf{49.4} & \textbf{58.5}  \\
\hline
GFLv2~\cite{gfocalv2}      & ResNet-101  & $\surd$          & 46.2 & 64.3 & 50.5 & 27.8 & 49.9 & 57.0  \\
DW+box refine (ours)   & ResNet-101  & $\surd$    & 46.8 & 65.1 & 50.5 & 27.7 & 49.9 & 59.1  \\
\hline
ATSS \cite{atss}       & ResNet-101-DCN  & $\times$     & 46.3 & 64.7 & 50.4 & 27.7 & 49.8 & 58.4  \\
PAA \cite{atss}        & ResNet-101-DCN  & $\times$     & 47.4 & 65.7 & 51.6 & 27.9 & 51.3 & 60.6  \\
GFL \cite{gfocal}      & ResNet-101-DCN  & $\times$     & 47.3 & 66.3 & 51.4 & 28.0 & 51.1 & 59.2  \\
MuSu \cite{musu}       & ResNet-101-DCN & $\times$      & 47.4 & 65.0 & 51.8 & 27.8 & 50.5 & 60.0  \\
VFL \cite{varifocalnet}        & ResNet-101-DCN & $\times$       & 48.5 & 67.4 & 52.9 & 29.1 & \textbf{52.2} & 61.9  \\
DW (ours)   & ResNet-101-DCN & $\times$         & \textbf{49.3} & \textbf{67.6} & \textbf{53.3} & \textbf{29.2} & \textbf{52.2} & \textbf{63.5}  \\
\hline
GFLv2 \cite{gfocalv2}      & ResNet-101-DCN & $\surd$     & 48.3 & 66.5 & 52.8 & 28.8 & 51.9 & 60.7  \\
DW+box refine (ours)  & ResNet-101-DCN & $\surd$    & 49.5 & 67.7 & 53.4 & 28.9 & 52.2 & 63.7  \\
\hline
ATSS \cite{atss}       & ResNeXt-101-64x4d & $\times$    & 45.6 & 64.6 & 49.7 & 28.5 & 48.9 & 55.6  \\
PAA \cite{paa}        & ResNeXt-101-64x4d & $\times$     & 46.6 & 65.6 & 50.8 & 28.8 & 50.4 & 57.9  \\
GFL \cite{gfocal}      & ResNeXt-101-64x4d & $\times$    & 46.0 & 65.1 & 50.1 & 28.2 & 49.6 & 56.0  \\
OTA \cite{ota}        & ResNeXt-101-64x4d & $\times$     & 47.0 & 65.8 & 51.1 & 29.2 & 50.4 & 57.9  \\
DW(ours)   & ResNeXt-101-64x4d   & $\times$     & \textbf{48.2} & \textbf{67.1} & \textbf{52.2} & \textbf{29.6} & \textbf{51.2} & \textbf{60.8}  \\
\hline
VFL~\cite{varifocalnet}        & ResNeXt-101-64x4d & $\surd$     & 48.5 & 67.0 & 52.6 & 30.1 & 51.7 & 59.7  \\
TOOD~\cite{tood}       & ResNeXt-101-64x4d & $\surd$     & 48.3 & 66.5 & 52.4 & \textbf{30.7} & 51.3 & 58.6 \\
DW+box refine (ours)  & ResNeXt-101-64x4d   & $\surd$ & 48.7 & 67.1 & 52.7 & 29.7 & 51.6 & 61.1  \\
DW$^*$ (ours) & ResNeXt-101-64x4d & $\surd$  & 49.8 & 67.7 & 53.8 & 30.4 & 52.3 & 63.0 \\
\bottomrule[1pt]
\end{tabular}
}
\end{table*}
\subsection{Comparison with State-of-the-Arts}
We compare our DW with other one-stage detectors on  \texttt{test-dev} 2017 in Table~\ref{comp_sota}. Following previous works~\cite{tood,varifocalnet,gfocal}, the multi-scale training strategy and 2$\times$ learning schedule (24 epochs) are adopted during training. We report the results of single-model single-scale testing for all methods. Other settings are consistent with \cite{tood,varifocalnet,gfocal}.

Apart from the LA strategy, some works~\cite{gfocalv2,tood,varifocalnet} also utilize additional feature learning modules to boost their detectors. For fair comparisons, in Table~\ref{comp_sota} we compare with them by reporting the performance w/wo this auxiliary module.
It can be seen that our DW method with ResNet-101 achieves 46.2 AP, outperforming all other competitive methods with the same backbone, including VFL (44.9 AP), GFL (45.0 AP) and OTA (45.3 AP). When using more powerful backbones like ResNet-101-DCN and ResNeXt-101-64x4d, DW reaches 49.3 and 48.2 AP, surpassing GFL by 2 and 2.2 points, respectively. We can also see that the operation of box refinement consistently improves DW with different backbones. It is worth mentioning that when we replace the detection head in FCOS by the one proposed in TOOD~\cite{tood}, DW achieves 49.8 AP, 1.5 points better than TOOD. This demonstrates the good generalization capacity of our DW strategy to other detection heads.
\begin{figure}[tbp]
    \centering
    \includegraphics[width=0.48\textwidth]{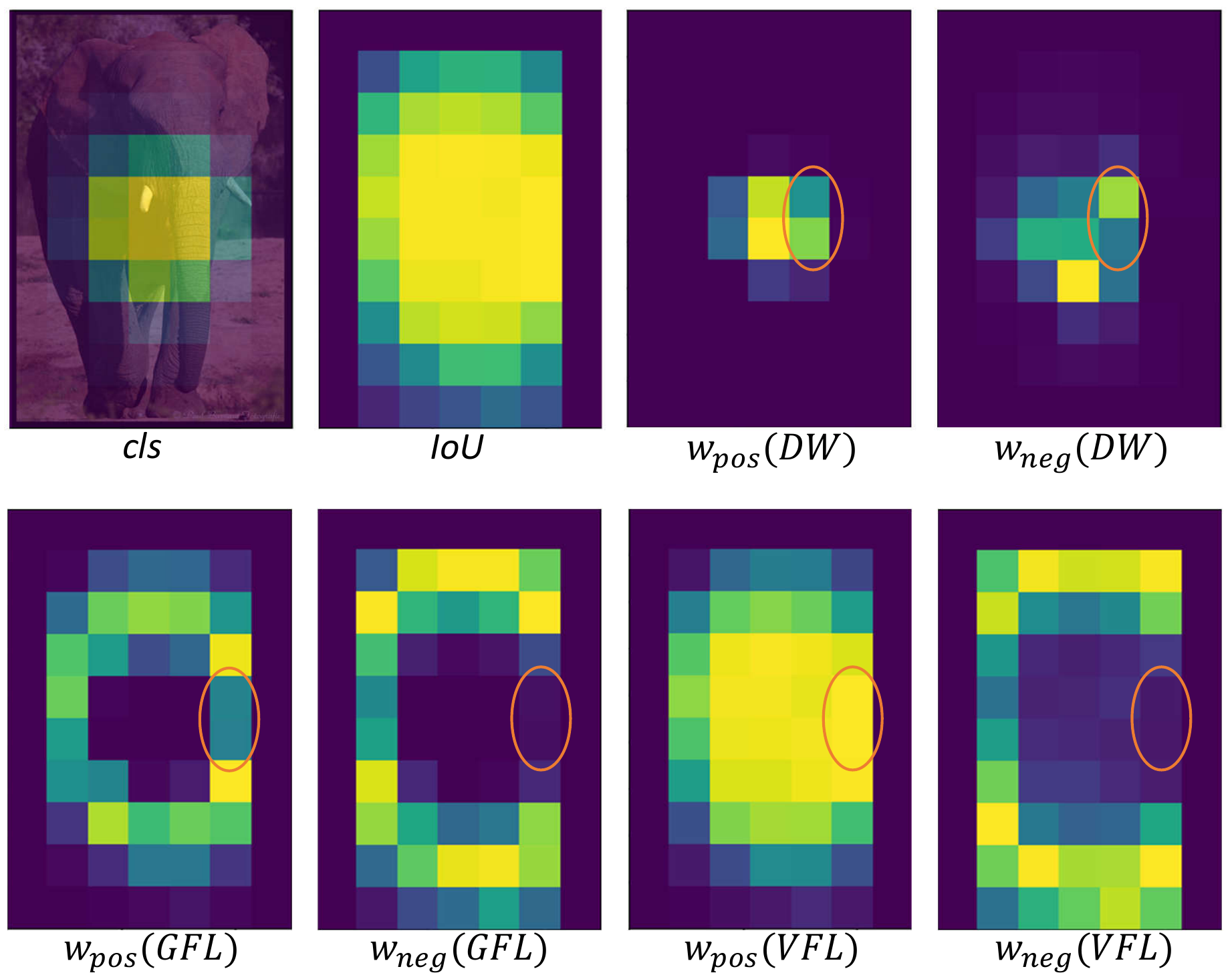}
    \caption{Visualization of \textit{cls} score, IoU, \textit{pos} and \textit{neg} weights.}
    \label{vis_AP}
\end{figure}
\subsection{Discussions}
\textbf{Visualization of DW.} To further understand the difference between DW and existing methods, we show the visualization maps of \textit{cls} score, IoU, \textit{pos} and \textit{neg} weights of DW and two representative methods, GFL~\cite{gfocal} and VFL~\cite{varifocalnet}, in Fig~\ref{vis_AP}. It can be seen that \textit{pos} and \textit{neg} weights in DW are mainly centralized on the central region of the GT, while GFL and VFL assign weights on a much wider region. This difference implies that DW can focus more on important samples and reduce the contribution of easy samples, such as those near the boundary of the object. This is why DW is more robust to the selection of candidate bag.

We can also see that anchors in the central region have distinct (\textit{pos}, \textit{neg}) weight pairs in DW. In contrast, the \textit{neg} weights are highly correlated with \textit{pos} weights in GFL and VFL. Anchors highlighted by the orange circle have almost the same \textit{pos} weight and \textit{neg} weight in GFL and VFL, while DW can distinguish them by assigning them different weights, providing the network higher learning capacity.

\textbf{Limitation of DW.} Although DW can well distinguish the importance of different anchors for an object, it will decrease the number of training samples at the same time, as shown in Fig~\ref{vis_AP}. This may affect the training efficacy on small objects. As shown in Table~\ref{comp_sota}, the improvement of DW on small objects is not as high as that on large objects. To mitigate this issue, we may dynamically set different hyper-parameters of $w_{pos}$ based on object size to balance the training samples between small and large objects.

\section{Conclusion}
We proposed an adaptive label assignment scheme, named dual weighting (DW), to train accurate dense object detectors. DW broke the convention of coupled weighting in previous dense detectors, and it dynamically assigned individual \textit{pos} and \textit{neg} weights for each anchor by estimating the consistency and inconsistency metrics from different aspects. A new box refinement operation was also developed to directly refine boxes on the regression map. DW was highly compatible with the evaluation metric. Experiments on the MS COCO benchmark verified the effectiveness of DW under various backbones. With and without box refinement, DW with ResNet-50 achieved 41.5 AP and 42.2 AP, respectively, recording new state-of-the-art. As a new label assignment strategy, DW also demonstrated good generalization performance to different detection heads.

Negative societal impacts of object detection mainly arise from the abuse on military applications and privacy issues, which warrants careful consideration before applying this technology to real life.

{\small
\bibliographystyle{ieee_fullname}
\bibliography{reference}
}

\end{document}